%% file: arxiv.tex
\documentclass[letterpaper]{article} % DO NOT CHANGE THIS
\usepackage{aaai24}  % DO NOT CHANGE THIS
\usepackage{times}  % DO NOT CHANGE THIS
\usepackage{helvet}  % DO NOT CHANGE THIS
\usepackage{courier}  % DO NOT CHANGE THIS
\usepackage[hyphens]{url}  % DO NOT CHANGE THIS
\usepackage{graphicx} % DO NOT CHANGE THIS
\usepackage{natbib}  % DO NOT CHANGE THIS AND DO NOT ADD ANY OPTIONS TO IT
\usepackage{caption} % DO NOT CHANGE THIS AND DO NOT ADD ANY OPTIONS TO IT
\usepackage{algorithm}
\usepackage{algorithmic}

\usepackage{newfloat}
\usepackage{listings}
\DeclareCaptionStyle{ruled}{labelfont=normalfont,labelsep=colon,strut=off} % DO NOT CHANGE THIS
\floatstyle{ruled}
\newfloat{listing}{tb}{lst}{}
\floatname{listing}{Listing}
\title{RegionBLIP: A Unified Multi-modal Pre-training Framework for Holistic and Regional Comprehension}
\author {
    Qiang Zhou,~ %\textsuperscript{\rm 1,\rm 2},
    Chaohui Yu,~ %\textsuperscript{\rm 2},
    Shaofeng Zhang,~ %\textsuperscript{\rm 1},
    Sitong Wu,~ %\textsuperscript{\rm 1},
    Zhibing Wang,~
    Fan Wang %\textsuperscript{\rm 1}
}
\affiliations {
    jianchong.zq@alibaba-inc.com,~~~huakun.ych@alibaba-inc.com,~~~sherrylone@sjtu.edu.cn
    wusitong98@gmail.com,~~~zhibin.waz@inftech.ai,~~~fan.w@alibaba-inc.com
}

\usepackage{bibentry}
\usepackage{booktabs}
\usepackage{multirow}
\usepackage{amsmath}
\usepackage{amssymb}
\usepackage{color}
\usepackage{comment}

\def\Ours{RegionBLIP}

\begin{document}

\maketitle

\begin{abstract}

\input{chapters/abstract}

\end{abstract}

\input{chapters/0_introduction}

\input{chapters/1_related}
\input{chapters/2_method}

\input{chapters/3_experiment}

\input{chapters/4_regioncap}
\input{chapters/5_conclusion}

\bibliography{aaai24}

\clearpage

\section{Appendix}
\input{chapters/region_text_data}

\end{document}

%% file: chapters/abstract.tex
In this work, we investigate extending the comprehension of Multi-modal Large Language Models (MLLMs) to regional objects.
To this end, we propose to extract features corresponding to regional objects as soft prompts for LLM, which provides a straightforward and scalable approach and eliminates the need for LLM fine-tuning. To effectively extract regional features from regular image features and irregular point cloud features, we present a novel and unified position-assisted feature extraction module. 
Furthermore, training an MLLM from scratch is highly time-consuming. Thus, we propose incrementally extending existing pre-trained MLLMs to comprehend more modalities and the regional objects of those modalities. Specifically, we freeze the Q-Former from BLIP-2, an impressive MLLM, and optimize the modality-specific Lora parameters in Q-Former and LLM for each newly introduced modality. The freezing of the Q-Former eliminates the need for extensive pre-training on massive image-text data. The freezed Q-Former pre-trained from massive image-text data is also beneficial for the pre-training on image-region-text data.
We name our framework \Ours{}.
We pre-train \Ours{} on image-region-text, point-cloud-text, and point-cloud-region-text data. Experimental results verify that \Ours{} can preserve the image comprehension capability of BILP-2 and further gain a comprehension of the newly introduced point cloud modality and regional objects.
The Data, Code, and Pre-trained models will be available at https://github.com/mightyzau/RegionBLIP.

%% file: chapters/0_introduction.tex
\section{Introduction}

%%% 1. Background
Large language models (LLM) like ChatGPT have demonstrated impressive text comprehension, reasoning, and generation capabilities, opening up opportunities for a wide range of practical applications. To further boost the data modalities and tasks that LLM can handle, recent research work, such as Flamingo~\cite{flamingo_AlayracDLMBHLMM22}, BLIP-2~\cite{blip2_2023}, etc., has introduced the image modality into LLM, setting off a research boom in multi-modal large language models (MLLM).

%%% 2. Incremental
%Recently, some works~\cite{} are proposed to pretrain with more modalities, e.g., video, voice etc., to make LLM comprehend much more modalities. However, with newly introduced modality and data, pre-training from scratch is much more time-consuming. To this end, we propose a incremental approach to extend existing MLLM to comprehend newly introduced modalities. BLIP-2 propose Q-Former to to extract text aligned features, which is to feed to a frozen LLM and is easily to be comprehended by the LLM. In this work, we freeze the Q-Former of BLIP-2, and additionally learn modality-speicfic Lora parameters in Q-Former and LLM for newly introduced point cloud modality and regional objects. 
In recent studies~\cite{maaz2023videochatgpt, zhang2023videollama, lyu2023macawllm,chen2023xllm,yin2023lamm}, efforts have been made to expand the comprehension capability of LLM to encompass a broader range of modalities, including image, video, voice, point clouds, and more. Typically, a new MLLM is pre-trained using data from all related modalities. However, this process of pre-training a new MLLM from scratch can be incredibly time-consuming. To address this issue, \textit{we propose an incremental approach for extending the existing MLLM (e.g., BLIP-2) to comprehend newly introduced modalities.} We freeze the Q-Former of BLIP-2 and further learn a set of modality-specific Lora~\cite{Lora_HuSWALWWC22} parameters in both Q-Former and LLM for the newly introduced point cloud modality and regional objects. In this way, we preserve the image comprehension capabilities of BLIP-2 and do not need to retrain our model on the massive image-text paired data used by BLIP-2.

In many application scenarios, such as virtual reality, it is more valuable to provide regional comprehension. 
Works such as Kosmos-2~\cite{kosmos2_abs-2306-14824} and Shikra~\cite{shikra_2306-15195} convert the location of regional objects into language descriptions as additional input to LLM. However, this scheme also has some limitations, such as the difficulty describing complex masks of regional objects and the need to fine-tune the LLM to understand the language position descriptions.
Our main motivation is that, \textit{extracting regional features is much simpler than making the LLM comprehend the regional position description}. To this end, we align regional features with text embeddings and feed the aligned regional features into an LLM for regional comprehension. RoIAlign~\cite{maskrcnn_HeGDG17} is a common strategy for extracting regional features in image modalities, but it is unsuitable for irregular point cloud features. 
In this work, we propose a unified position-assisted feature extraction (PaFE) module, which can effectively extract regional features from regular image features and irregular point cloud features for LLM comprehension.

%To enhance \Ours's comprehension at the regional level, we propose an automated regional data mining pipeline and extract more than 10M image-region-text paired data (RegionCap-10M) from extensive image sources including CC3M~\cite{cc3m_SoricutDSG18} and CC12M~\cite{cc12m_ChangpinyoSDS21}. Due to this data mining effort, our pre-trained \Ours{} has outstanding region captioning abilities. Notably, our model achieves remarkable zero-shot box captioning performance on the RefCOCO dataset, achieving the CIDEr score of xx.

Compared with image-text data, the dataset size of image-region-text data is still relatively small. Therefore, we design a mining approach for image-region-text data and generate the \textit{RegionCap-10M} dataset. We expect this larger-scale dataset to extend the capability of LLM to comprehend image regions in various scenes.

In summary, our contributions are as follows.
\begin{itemize}
    \item We propose a simple and unified scheme for LLM to comprehend regional objects in image and point cloud modalities.
    \item We propose an incremental pre-training scheme that optimizes only the modality-specific Lora parameters in Q-Former and LLM, which extends BLIP-2's comprehended modalities expeditiously and effectively.

    \item We will release the RegionCap-10M dataset to help improve the capability of LLM to comprehend image regions in open scenes.
\end{itemize}

%% file: chapters/1_related.tex
\section{Related Work}

\subsection{Multi-modal Large Language Model}

Research and development of multi-modal versions of large language models has attracted the interest of many researchers and practitioners in this field.

One line of research is to employ multi-modal data to fine-tune the LLM. 
Flamingo~\cite{flamingo_AlayracDLMBHLMM22} converts image or video features of various sizes into a fixed number of visual outputs through the proposed perceiver resampler module as the input of LLM.
To condition the LLM on visual inputs, Flamingo inserts new cross-attention layers between existing LLM layers, and trains on a large-scale interleaved image-text dataset.
OpenFlamingo~\cite{openflanmingo_anas_awadalla_2023_7733589} is a reimplementation of Flamingo that is open-sourced to the community.
MM-GPT~\cite{gong2023multimodalgpt} fine-tunes OpenFlamingo by adding Lora parameters to the LLM model, and achieves more user-friendly interactions by carefully constructing instruction datasets.
Otter~\cite{li2023otter} is also fine-tuned on OpenFlamingo, and it demonstrates improved instruction-following and in-context learning capabilities through its carefully constructed multi-modal in-context instruction tuning (MIMIC-IT) dataset.

Another line of research is to align modalities like image to text that LLM can comprehend.
BLIP-2~\cite{blip2_2023} pre-trains in two stages. In stage 1, the proposed Q-Former is used to extract a fixed number of query features from various size image features, and these query features are aligned with the text through multiple vision-language losses. In stage 2, these aligned query features are fed to a frozen LLM as soft visual prompts and are pre-trained with language modeling loss.
Mini-GPT4~\cite{zhu2023minigpt4}, mPLUG-OWL~\cite{ye2023mplugowl}, VPGTrans~\cite{zhang2023transfer}, and InstructBLIP~\cite{dai2023instructblip} retain the Q-Former design of BLIP-2, replace the language model with a larger one, and fine-tune on carefully collected instruction data. 

Instead of instructing fine-tuning, we focus on multi-modal pre-training to enhance LLM's capability to comprehend a broader range of modalities. We propose an incremental pre-training framework, which is much more training-efficient.

\subsection{MLLM for Regional Comprehension}

The comprehension of regional objects has garnered significant focus and investigation in the context of unified visual-language pre-training frameworks, as well as in the more recent development of MLLM.

VL-T5~\cite{VL-T5_ChoLTB21} converts the visual grounding task into regional feature conditioned text generation, in which the regional feature is encoded as a sum of RoI (Region of Interest) features, RoI coordinates, image id, and region id.
OFA~\cite{ofa_WangYMLBLMZZY22} converts continuous corner coordinates of regional objects to location tokens and pre-train in a unified sequence-to-sequence abstraction via handcrafted instructions.
Similarly, PEVL~\cite{PEVL_YaoCZ0LCS22} reformulates object positions as discrete tokens, and learns the joint distribution of object positions and language in a unified language modeling framework.
Recently, in the field of MLLM, Shikra~\cite{shikra_2306-15195} handles spatial coordinate inputs and outputs in natural language without introducing extra vocabulary or position encoders. During training, both the modal adapter layer and the entire LLM are optimized.
Kosmos-2~\cite{kosmos2_abs-2306-14824} constructs a web-scale grounded image-text dataset, and the object location descriptions are converted to sequences of location tokens during training. Kosmos-2 also needs to optimize the entire LLM model to comprehend the input visual location tokens.

In this work, instead of fine-tuning LLM to comprehend regional location descriptions (language or discrete tokens), we feed the text-aligned regional features to the frozen LLM.
We propose a unified position-assisted regional feature extraction module, which can extract regional features from regular image features and irregular point cloud features.

%% file: chapters/2_method.tex
\section{Method}

In this section, we introduce \Ours{}, an unified multi-modal pre-training framework permits regional comprehension of image and point cloud modalities.
Specifically, in Section~\ref{sec:model_structure}, we provide an outline of the model architecture. 
Section~\ref{sec:model_pretraining} covers the details of model pre-training.
%Section~\ref{sec:box_text_dataset} outlines the process of gathering image-region-text data from a vast pool of image-text data. 

\begin{figure*}[!t]
    \centering
    \includegraphics[width=0.95\textwidth]{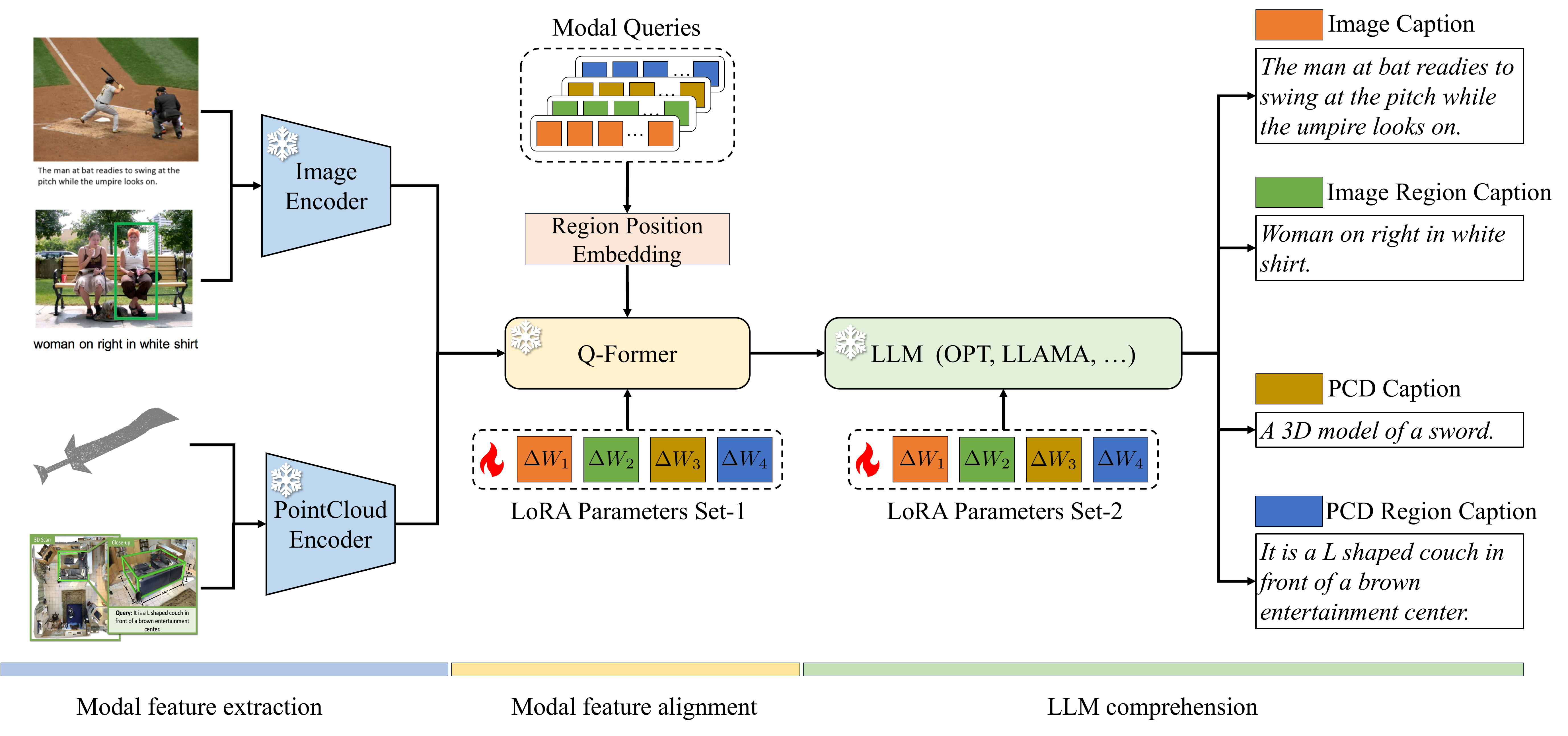}
    \caption{\Ours{} is a unified incremental pre-training framework supporting LLM's comprehension of images, point clouds, and regional objects.
    For efficient pre-training, \Ours{} freezes the Q-Former of BLIP-2~\cite{blip2_2023} and learns a set of modality-specific Lora parameters for newly added modalities.
    To effectively extract region features from regular image features and irregular point cloud features, \Ours{} proposes a unified scheme of position-assisted region feature extraction module.
    }
    \label{fig:model_structure}
\end{figure*}

\subsection{Model Architecture}
\label{sec:model_structure}

As shown in Figure~\ref{fig:model_structure},  \Ours{} has three primary modules: modal feature extraction, modal feature alignment, and LLM comprehension.
The feature extraction module extracts features from different modalities, such as images (I) and point clouds (P). The feature alignment module aligns these modality features with text embeddings to facilitate the LLM's comprehension of these modality inputs. The final frozen LLM model processes these aligned modality features as input to generate the final text comprehension.

\textbf{Modal feature extraction.} \Ours{} aims to comprehend image and point cloud modalities and their regional objects. However, we defer extracting fine-grained regional features to the subsequent modal feature alignment module. In contrast, we only extract the overall features of the image or point cloud input here. In this way, we can share image encoders in I-text and I-region-text data and point cloud encoders in P-text and P-region-text data. We employ the pre-trained CLIP~\cite{CLIP_RadfordKHRGASAM21,eva_clip_abs-2211-07636} model's vision encoder as the image encoder and freeze its parameters during our model pre-training. We employ the pre-trained Point-BERT~\cite{pointbert_YuTR00L22} model as the point cloud encoder and also freeze its parameters.

\textbf{Modal feature alignment.} LLM is trained with language corpus, making comprehending image or point cloud inputs challenging. Aligning image or point cloud features with textual descriptions before feeding them into LLM will facilitate subsequent LLM comprehension. In this work, we utilize several learnable queries (32 by default) to extract the aligned fixed-length image or point cloud features from the output of the image encoder or point cloud encoder. Specifically, we use the Q-Former proposed in BLIP-2~\cite{blip2_2023} to extract fixed-length aligned image or point-cloud features.
The Q-Former is shared between the I-text and I-region-text data and between the P-text and P-region-text data.
Instead of retraining Q-Former with massive image-text data, we freeze Q-Former's parameters from the pre-trained BLIP-2 model to preserve its image comprehension capability. 
The frozen Q-Former pre-trained on image-text data performs poorly on regional objects. To this end, we propose learning a set of modality-specific Lora parameters in Q-Former for modalities of image, point cloud and their regional objects. Learnable queries are also not shared between modalities.
In such an incremental pre-training scheme, \Ours{} can build on the existing image comprehension capabilities of BLIP-2 while rapidly extending to comprehend more modalities.

%\textcolor{blue}{TO REMOVE:
%To simplify the model structure, we adopt a shared Q-Former for all modality inputs, including I-text, I-region-text, P-text, and P-region-text. However, our experiments show that a shared Q-Former leads to conflicts between modalities during mixed training, leading to slower model convergence and poor performance. To address this, we introduced an additional set of Lora parameters for each modality within Q-Former, as shown in Figure~\ref{fig:model_structure}. Learnable queries are also not shared between modalities. Experimental results in Table~\ref{xxx} show that this approach significantly improves model convergence and performance on various modal tasks.}

For LLM to comprehend regional objects in images, existing methods such as Kosmos-2~\cite{kosmos2_abs-2306-14824} and Shikra~\cite{shikra_2306-15195} convert the position of regional objects into language descriptions and then finetune LLM to comprehend the language position descriptions.
Differently, we take a more straightforward, unified approach to support LLM's comprehension of regional objects. Similar to feeding image features into LLM for image comprehension, we directly feed features of regional objects into LLM for region-level understanding. Our experimental results show that in this way, LLM achieves promising regional object comprehension capability without finetuning. 
To effectively extract regional features, we propose a unified position-assisted feature extraction (PaFE) scheme, which is suitable for regular image features and irregular point cloud features. Specifically, we transform the normalized position coordinates of region objects into position embeddings via a two-layer MLP network. The position embedding is then added to the learnable modal queries to help extract regional features via the shared Q-Former network, as shown in Figure~\ref{fig:model_structure}.

\textbf{LLM comprehension.} We feed text-aligned features from Q-Former into a frozen LLM to utilize the generative language capabilities of LLM. Likewise, we also learn a set of Lora parameters for each modality in LLM to mitigate conflicts among various modalities.
We experimented with two types of LLM: decoder-based LLM and encoder-decoder-based LLM. For decoder-based LLMs (e.g., OPT~\cite{OPT_abs-2205-01068}), we use a language modeling loss for pre-training, where the task of the frozen LLM is to generate text conditioned on the extracted modality features of the Q-Former. 
We pre-train with a prefix language modeling loss for encoder-decoder-based LLMs (e.g., FlanT5~\cite{FlanT5_abs-2210-11416}), using a fixed prefix of ``a photo of'' for image modality and ``a point cloud of'' for point cloud modality. 
The prefix text is concatenated with the modality features of Q-Former as input to the LLM encoder. The suffix text is employed as the generation target for the LLM decoder.
Furthermore, we employ a regression auxiliary loss $\mathcal{L}_{reg}$ to enhance the extraction of regional features. Specifically, we utilize the regional features obtained from Q-Former as input, employ a fully connected layer to predict regional objects' normalized coordinates and utilize L1 loss as the optimization objective.
\begin{equation}
    \mathcal{L}_{reg} = L_1(p, p^*),
\end{equation}
where $p$ is the predicted regional object's normalized coordinates and $p^*$ is the corresponding ground-truth values.

\subsection{Model Pre-training}
\label{sec:model_pretraining}

\paragraph{Pre-training strategy.}
\Ours{} combines data from various modalities (including I-text, I-region-text, P-text, and P-region-text) and adopts a single-stage pre-training strategy, which will be illustrated in the following.

\textit{Joint pre-training of modality alignment and LLM comprehension.}
BLIP-2's pre-training process consists of two stages, i.e., modality alignment and LLM comprehension. In the modality alignment stage, the visual features of the frozen image encoder are aligned with the text features. In the LLM comprehension stage, the frozen LLM is utilized to generate text output by taking the aligned visual features as input. 
Our approach differs in that we employ a single-stage pre-training approach consisting of all pre-training losses in BLIP-2~\cite{blip2_2023}:
\begin{equation}
    \mathcal{L} = \mathcal{L}_{ITC} + \mathcal{L}_{ITG} + \mathcal{L}_{ITM} + \mathcal{L}_{LLM} + 
    \lambda \cdot \mathcal{L}_{reg},
\end{equation}
where $\mathcal{L}_{ITC}$ is the I/P-text contrastive learning loss , $\mathcal{L}_{ITG}$ is the I/P-grounded text generation loss, and $\mathcal{L}_{ITM}$ is the I/P-text matching loss.
$\mathcal{L}_{LLM}$ is the language modeling loss for LLM comprehension. 
$\mathcal{L}_{reg}$ is the auxiliary regression loss, and its loss weight $\lambda$ is set to 1.0 by default.
Our experiments demonstrate that the one-stage pre-training strategy can significantly improve pre-training efficiency without compromising model performance.

\textit{Multi-modal semi-hybrid pre-training.}
\Ours{}'s pre-training data covers a variety of modalities, including I-text, I-region-text, P-text, and P-region-text. Each modality has its distinct inference operation. For instance, P-text and P-region-text data necessitate a point cloud encoder rather than an image encoder. Furthermore, I-region-text or P-region-text data requires additional position embedding computation. Due to this, training on mixed multi-modal data efficiently is complicated. To this end, we conduct a semi-hybrid pre-training approach. We mix data from all modalities for model pre-training and sequentially pre-train each modality's data at each epoch.

\paragraph{Incremental pre-training setting.}
Compared with the I-region-text, P-text, and P-region-text data used in this study, the amount of I-text data used by BLIP-2 is considerable, reaching 129M. Incorporating such a large amount of I-text data directly into the training process will significantly extend the training period. To this end, we propose a fast incremental pre-training scheme that eliminates the necessity of I-text data.
Specifically, we import and freeze the Q-Former parameters from the pre-trained BLIP-2 model and optimize the corresponding Lora parameters only for I-region-text modality. As shown in Figure~\ref{fig:model_structure}, each modality has a set of learnable Lora parameters in Q-Former and LLM.
This incremental pre-training scheme enables us to inherit image comprehension capabilities from the pre-trained BLIP-2 model while gaining comprehension abilities for newly added modalities through Lora parameter optimization. Overall, this training scheme provides a cost-effective solution for LLM to integrate more modal comprehension capabilities.

\paragraph{Pre-training data.}
In this work, we develop a framework that enables LLM to comprehend data from two modalities, namely images and point clouds. Furthermore, we support global understanding and regional object understanding for each modality. Specifically, the training data for each modality used is as follows.

\textit{I-text.} 
%We utilize the same pre-training datasets as BLIP~\cite{blip_0001LXH22}, which encompasses a total of 129M images, including those from COCO~\cite{coco_LinMBHPRDZ14}, Visual Genome~\cite{vg_dataset_Krishna2016VisualGC}, CC3M~\cite{cc3m_SoricutDSG18}, CC12M~\cite{cc12m_ChangpinyoSDS21}, SBU~\cite{sbu_OrdonezKB11}, and 115M images from the LAION400M~\cite{laion400m_abs-2111-02114} dataset.
As stated in the incremental pre-training setting, the training of our \Ours{} does not involve image-text paired data.

\textit{I-region-text.}
We train our model using the RefCOCO~\cite{refcoco} training set, which encompasses 113k paired data of box and text. 
We evaluate our model on the RefCOCO test set. 
%To enhance the LLM's proficiency in comprehending image regions within more scenes, we extract more than 10M pairs of box and text data (\textbf{RegionCap-10M}) from image-text datasets of CC3M~\cite{cc3m_SoricutDSG18}, CC12M~\cite{cc12m_ChangpinyoSDS21}, and COYO-700M~\cite{coyo700m}. 
%Further details regarding this extraction process can be found in Section~\ref{sec:box_text_dataset}. 
%We intend to open-source the 10M I-region-text data, and the automatic extraction code will also be made public.

\textit{P-text.}
We use the newly released Objaverse~\cite{objaverse_abs-2212-08051} as the point cloud data source and obtain the corresponding captions from Cap3D~\cite{cap3d_abs-2306-07279}. The corpus contains approximately 660K point-cloud-text paired data. We randomly select 20K samples as the test set, and the rest of the data is retained for training.

\textit{P-region-text.}
For the paired data of point-cloud-region-text, we utilize the ScanRefer~\cite{scanrefer_ChenCN20} dataset, a large scale dataset containing 51,583 descriptions of 11,046 3D objects from 800 ScanNet~\cite{scannet_DaiCSHFN17} scenes.

\paragraph{Pre-training hyperparameters.}

All models are pre-trained on the 8$\times$A100 machine and use the same pre-training hyperparameters.
We use the AdamW~\cite{adamw_LoshchilovH19} optimizer with $\beta_1 = 0.9$, $\beta_2 = 0.999$, and a weight decay of 0.05. We use a cosine learning rate decay with a peak learning rate of 1e-4 and a linear warmup of 200 steps. The minimum learning rate is set to 1e-5. 
We use images of size $224 \times 224$ for image-text and image-region-text data.
For point clouds, we set the number of point groups in Point-BERT~\cite{pointbert_YuTR00L22} to 512 and use augmentation with random dropout, scaling, and rotation.

%% file: chapters/3_experiment.tex
\begin{figure*}[!t]
    \centering
    \includegraphics[width=1.0\linewidth]{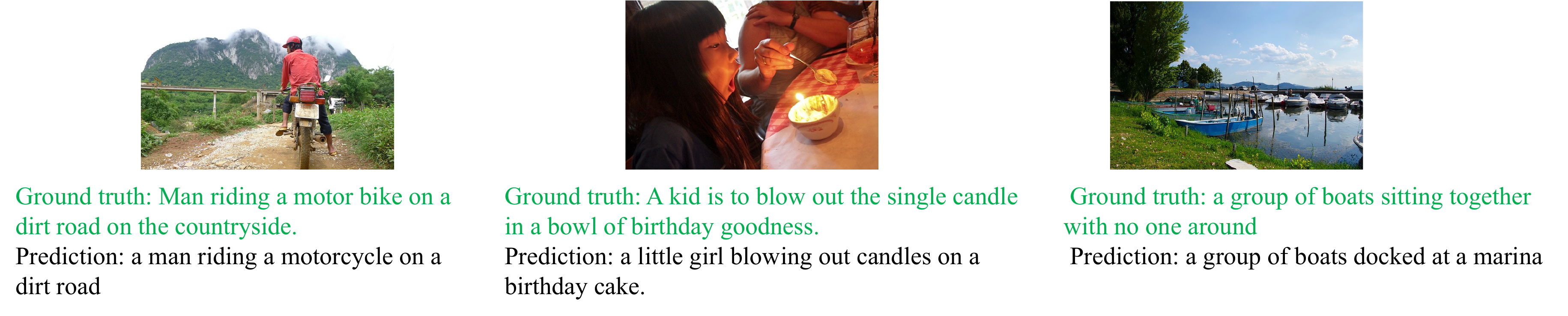}
    \caption{Examples of image captioning. The samples are from the COCO caption~\cite{coco_LinMBHPRDZ14} test set, and the model is \Ours{} OPT$_{2.7B}$.}
    %\vspace{-0.1em}
    \label{fig:example_img}
\end{figure*}

\begin{figure*}[!t]
    \centering
    \includegraphics[width=1.0\linewidth]{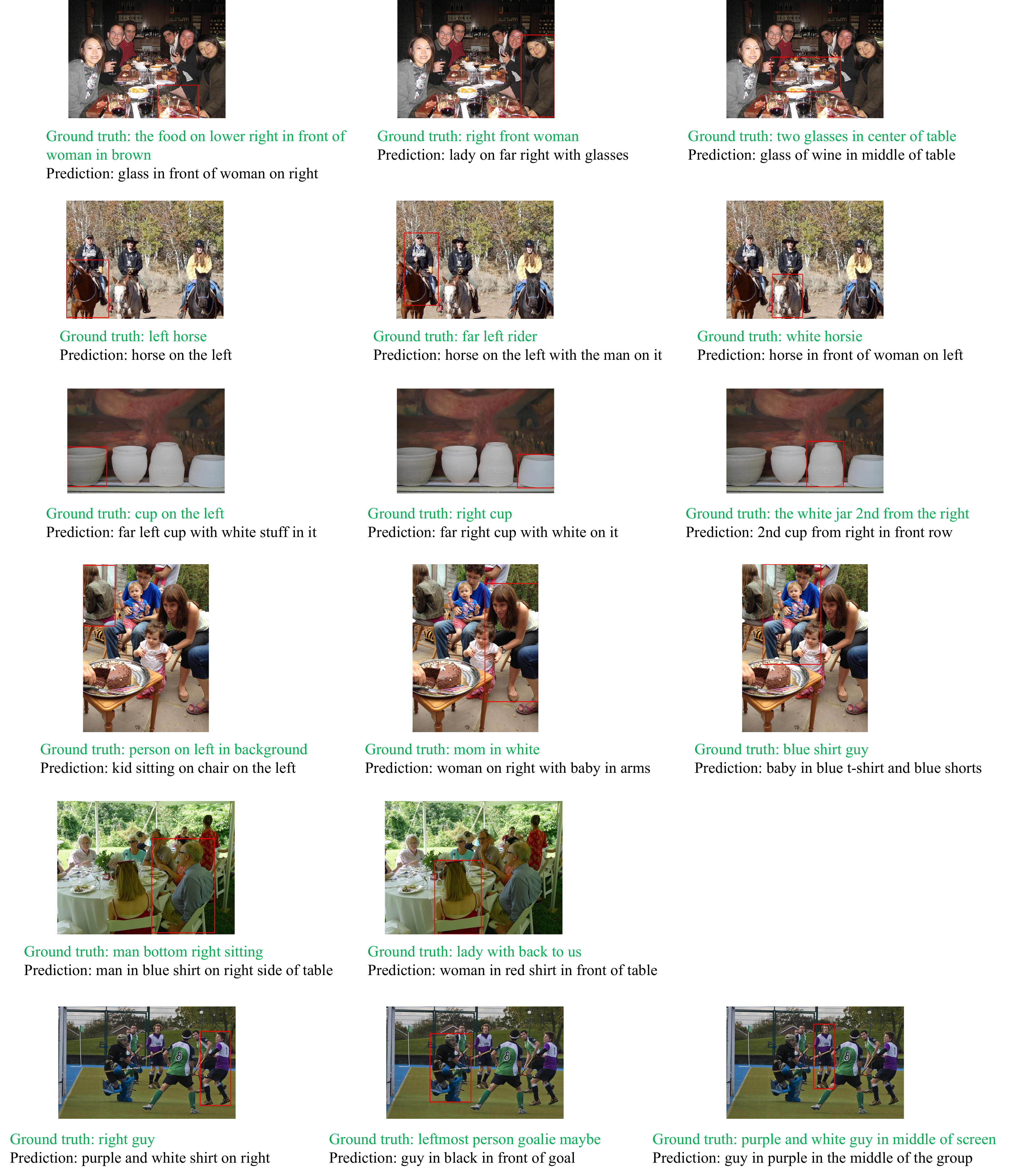}
    \caption{Examples of image-region captioning. The samples are from the RefCOCO~\cite{refcoco} test set, and the model is \Ours{} OPT$_{2.7B}$.}
    \label{fig:example_img_region}
\end{figure*}

\begin{figure*}[!t]
    \centering
    \includegraphics[width=0.82\linewidth]{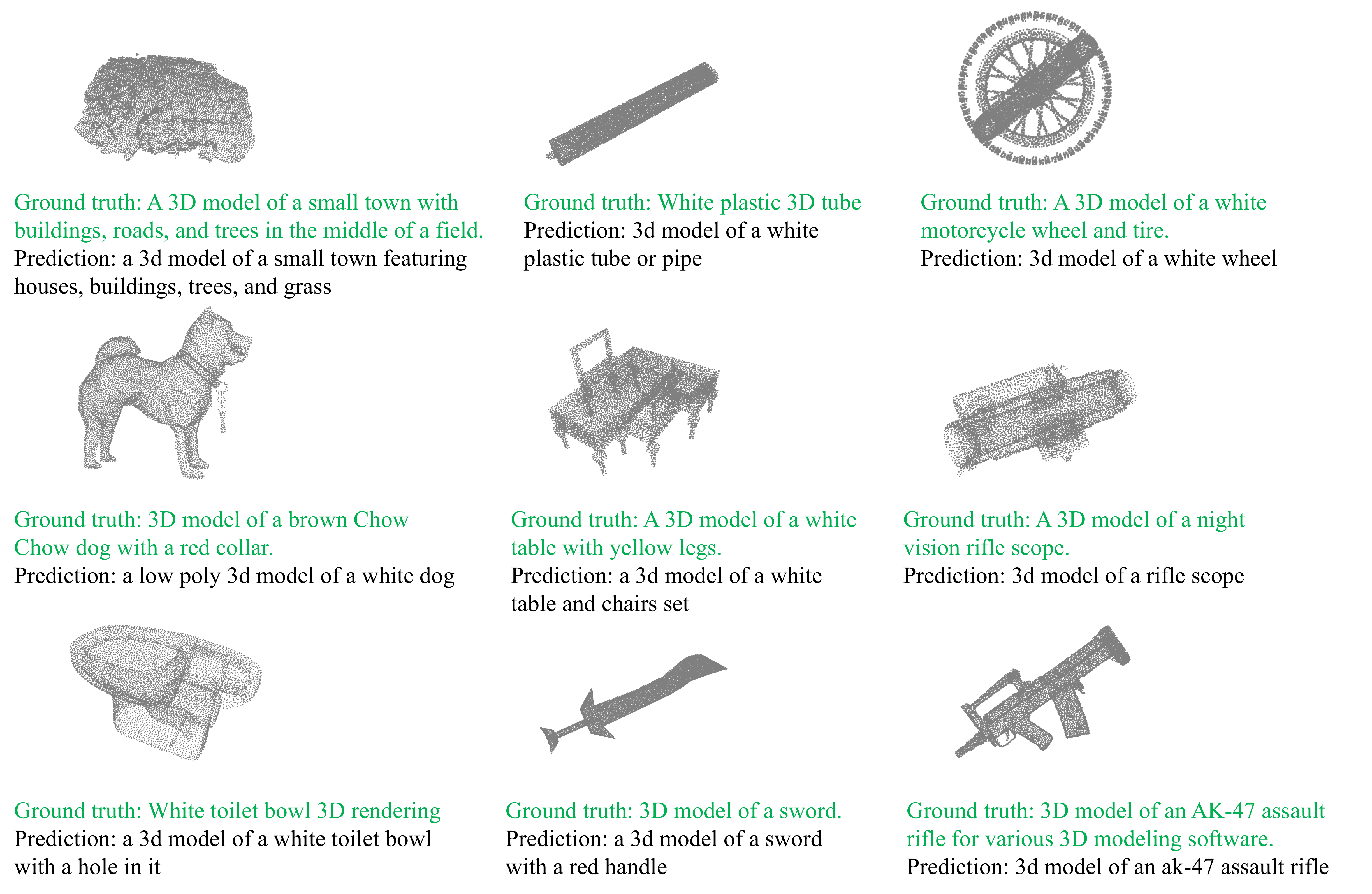}
    \caption{Examples of point cloud captioning. The samples are from the Objaverse~\cite{objaverse_abs-2212-08051} test set, and the model is \Ours{} OPT$_{2.7B}$.}
    \label{fig:example_pc}
\end{figure*}

\begin{figure*}[!t]
    \centering
    \includegraphics[width=0.81\linewidth]{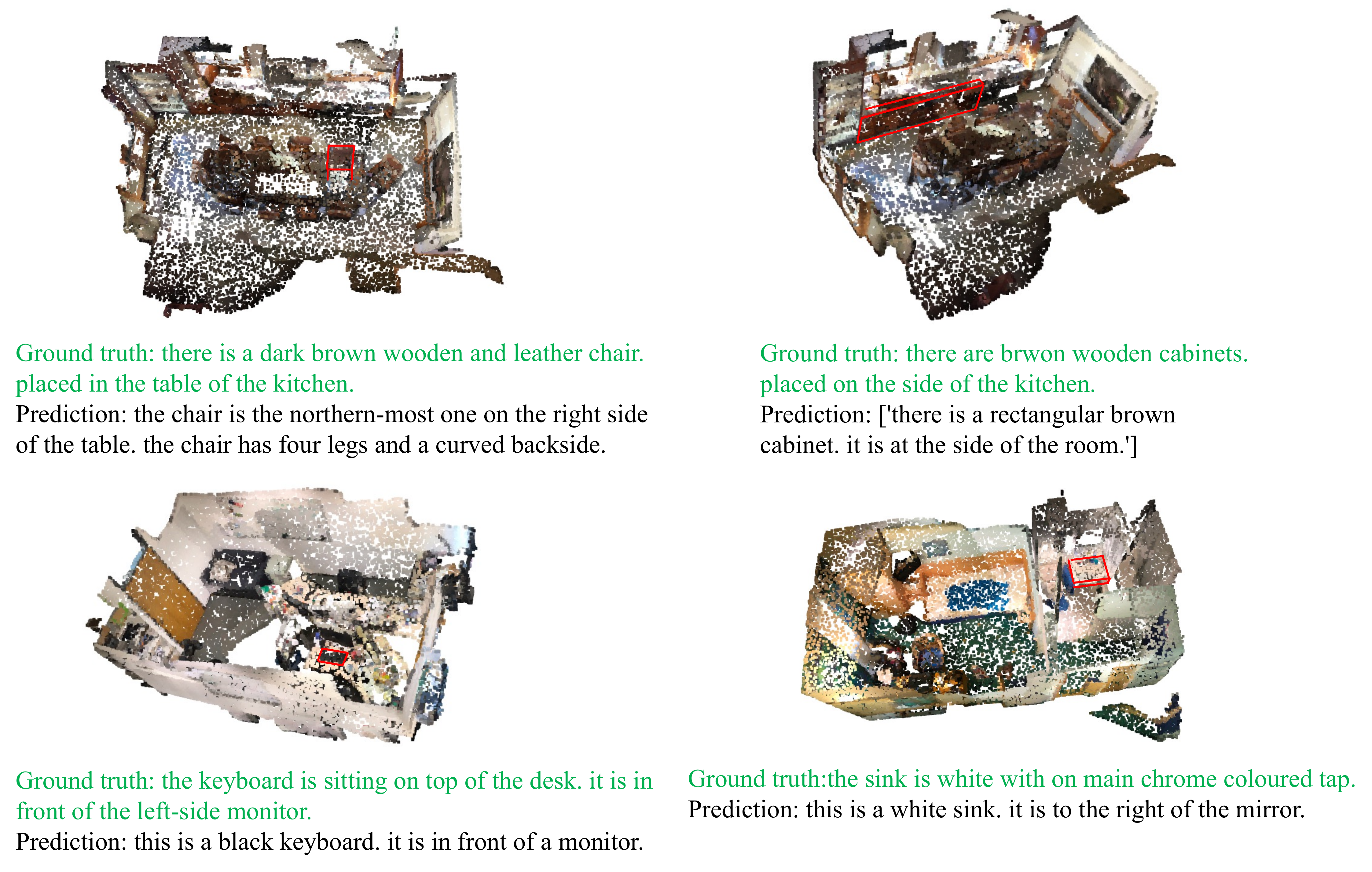}
    \caption{Examples of point-cloud-region captioning. The samples are from the ScanRefer~\cite{scanrefer_ChenCN20} validation set, and the model is \Ours{} OPT$_{2.7B}$.
    In this work, we did not utilize the color information of the point cloud, which limits the performance of point cloud region captioning to some extent.}
    \label{fig:example_pc_region}
\end{figure*}

\section{Experiment}

In this section, we experimentally investigate the comprehension performance on modalities of image, point cloud, and regional objects. 
We also present visualization examples of model comprehension on images (Figure~\ref{fig:example_img}), point clouds (Figure~\ref{fig:example_pc}), image regions (Figure~\ref{fig:example_img_region}), and pint cloud regions (Figure~\ref{fig:example_pc_region}), respectively.
Especially in the image region comprehending example shown in Figure~\ref{fig:example_img_region}, our model can accurately caption the specified regional objects, which demonstrates the effectiveness of our proposed PaFE module.

\subsection{Overall Performance}

\begin{table*}[!t]
\centering
\resizebox{1.0\linewidth}{!}{
\begin{tabular}{@{}c|c|ccc|cc|cc|cc@{}}
\toprule
\multirow{3}{*}{Models} & \multirow{3}{*}{Pre-train setting} & \multicolumn{3}{c|}{\textbf{Image Captioning}}                                                  & \multicolumn{2}{c|}{\textbf{Image-region Captioning }} & \multicolumn{2}{c|}{\textbf{PCD Captioning}}        & \multicolumn{2}{c}{\textbf{PCD-region Captioning}} \\ 
                &        & \multicolumn{1}{c|}{VQAv2 (val)} & \multicolumn{2}{c|}{COCO (Karpathy test)} & \multicolumn{2}{c|}{RefCOCO (test)}     & \multicolumn{2}{c|}{Objaverse (test)} & \multicolumn{2}{c}{ReferScannet (val)}     \\
                &        & \multicolumn{1}{c|}{VQA acc.}         & \multicolumn{1}{c}{CIDEr}     & SPICE    & \multicolumn{1}{c}{CIDEr}    & SPICE   & \multicolumn{1}{c}{CIDEr}   & SPICE  & \multicolumn{1}{c}{CIDEr}  & SPICE  \\ \midrule
                        
BLIP-2 OPT$_{2.7B}$      &              & \multicolumn{1}{c|}{51.88}                 & \multicolumn{1}{c}{130.7}          &    23.8      & \multicolumn{1}{c}{15.1}         &   13.2      & \multicolumn{1}{c}{-}        &  -      & \multicolumn{1}{c}{-}       &   -     \\
BLIP-2 OPT$_{6.7B}$     &              & \multicolumn{1}{c|}{54.01}                 & \multicolumn{1}{c}{130.3}          &  23.7        & \multicolumn{1}{c}{13.9}         &   12.6      & \multicolumn{1}{c}{-}        &   -     & \multicolumn{1}{c}{-}       &      -  \\
BLIP-2 Flant5-xl     &              & \multicolumn{1}{c|}{63.13}                 & \multicolumn{1}{c}{123.0}          &  22.0       & \multicolumn{1}{c}{19.6}         &   13.7      & \multicolumn{1}{c}{-}        &   -     & \multicolumn{1}{c}{-}       &      -  \\
BLIP-2 Flant5-xxl     &              & \multicolumn{1}{c|}{65.11}                 & \multicolumn{1}{c}{117.8}          &  21.1       & \multicolumn{1}{c}{22.3}         &   14.4      & \multicolumn{1}{c}{-}        &   -     & \multicolumn{1}{c}{-}       &      -  \\
\midrule

\Ours{} OPT$_{2.7B}$ (\textbf{ours}) &    \multirow{4}{*}{Freeze Q-Former}       & \multicolumn{1}{c|}{51.88}                 & \multicolumn{1}{c}{130.7}          &    23.8      & \multicolumn{1}{c}{63.5}         &  21.3       & \multicolumn{1}{c}{112.7}        &  31.6      & \multicolumn{1}{c}{57.0}       &  13.5      \\ 
\Ours{} OPT$_{6.7B}$ (\textbf{ours})  &          & \multicolumn{1}{c|}{54.01}                 & \multicolumn{1}{c}{130.3}          &    23.7      & \multicolumn{1}{c}{\textbf{64.2}}         &  20.9       & \multicolumn{1}{c}{\textbf{113.6}}        &  31.6      & \multicolumn{1}{c}{\textbf{59.3}}       &  14.4      \\ 

\Ours{} Flant5-xl (\textbf{ours})  &          & \multicolumn{1}{c|}{63.13}                 & \multicolumn{1}{c}{123.0}          &    22.0      & \multicolumn{1}{c}{47.6}         &  17.8       & \multicolumn{1}{c}{108.1}        &  31.7      & \multicolumn{1}{c}{59.2}       &  13.5      \\ 
\Ours{} Flant5-xxl (\textbf{ours})  &          & \multicolumn{1}{c|}{65.11}                 & \multicolumn{1}{c}{117.8}          &    21.1      & \multicolumn{1}{c}{56.1}         &  18.4       & \multicolumn{1}{c}{109.0}        &   31.5     & \multicolumn{1}{c}{53.6}       &    13.0   \\

\bottomrule
\end{tabular}
}
\caption{Overview of \Ours{}'s comprehension performance for various modalities.
Compared to BLIP-2~\cite{blip2_2023}, \Ours{} extends comprehension to more modalities by learning a set of Lora parameters for each modality in Q-Former and LLM. \Ours{} also extends the comprehension of regional objects with position-assisted feature extraction (PaFE) module.
Note that when testing the captioning performance of BLIP-2 on regional objects, we generate captions on cropped images by utilizing the given box coordinates.
}
\label{tbl:overall}
\end{table*}

To quantitatively evaluate RegionBLIP's capability to comprehend various modalities, we report RegionBLIP's captioning performance on images, image regions, point clouds, and point cloud regions, as shown in Table~\ref{tbl:overall}.
Thanks to the freezing of Q-Former, \Ours{} is able to preserve the image comprehension capabilities of BLIP-2, as evidenced by the results presented in Table~\ref{tbl:overall}, where \Ours{} achieves similar performance to that of BLIP-2 on both VQAv2 and COCO captioning tasks. 
%In addition, the freezing of Q-Former from BLIP-2 eliminates the need for extensive image-text data for pre-training, resulting in significantly higher training efficiency. 
By optimizing the Lora parameters in Q-Former and LLM, \Ours{} is capable of extending the capabilities of BLIP-2 to point cloud and region comprehension while utilizing only I-region-text, P-text, and P-region-text data.
Specifically, when taking the OPT$_{6.7B}$ as the LLM, \Ours{} achieves a CIDEr score of 64.2 on the RefCOCO~\cite{refcoco} test set, 113.6 on the Objavese test set, and 59.3 on the ReferScannet~\cite{scanrefer_ChenCN20} validation set. These results substantiate the efficacy of the proposed framework and pre-training setting of \Ours{}, which have enabled the swift expansion of MLLM's capability to comprehend a broader range of modalities.

%\subsection{Ablations}

\begin{table}[!t]
\centering
\resizebox{1.0\linewidth}{!}{
\begin{tabular}{@{}c|cc|cc|cc@{}}
\toprule
\multirow{3}{*}{PaFE} &          \multicolumn{2}{c|}{\textbf{Image-region Captioning}} & \multicolumn{2}{c|}{\textbf{PCD Captioning}}        & \multicolumn{2}{c}{\textbf{PCD-region Captioning}} \\
                      &  \multicolumn{2}{c|}{RefCOCO (test)}     & \multicolumn{2}{c|}{Objaverse (test)} & \multicolumn{2}{c}{ReferScannet (val)}     \\ 
                      
     &      \multicolumn{1}{c}{CIDEr}    & SPICE   & \multicolumn{1}{c}{CIDEr}   & SPICE  & \multicolumn{1}{c}{CIDEr}  & SPICE  \\ \midrule
                      
    & \multicolumn{1}{c}{44.7}         &    14.8     & \multicolumn{1}{c}{84.7}        &  27.6      & \multicolumn{1}{c}{21.4}       &   8.4     \\
                      
    $\checkmark$       & \multicolumn{1}{c}{\textbf{63.5}}         &    21.3     & \multicolumn{1}{c}{\textbf{112.7}}     &   31.6     & \multicolumn{1}{c}{\textbf{57.0}}       &    13.5    \\ \bottomrule
\end{tabular}
}
\caption{The impact of PaFE on the model's regional comprehension performance. 
The \textit{\Ours{}~OPT$_{2.7B}$} model was employed in this study.}
\label{tbl:ablation_PaFE}
\end{table}

\subsection{Impact of PaFE on Regional Comprehension}
Position-assisted feature extraction (PaFE) is proposed to extract regional features from images or point clouds. In this section, we conduct experiments to assess the efficacy of PaFE in image region captioning and point cloud region captioning tasks. The results in Table~\ref{tbl:ablation_PaFE} indicate that the model employing PaFE performs significantly better concerning CIDEr scores on the RefCOCO test set and ReferScannet validation set in comparison to the model that does not employ PaFE. 
Without PaFE, the model would confuse point-cloud caption and point-cloud-region caption tasks. Thus, the CIDEr score on the Objaverse test set is also affected, decreasing from 112.7 to 84.7.
These results validate the effectiveness of our proposed PaFE module in enhancing regional comprehension of LLM.

%% file: chapters/4_regioncap.tex
\section{RegionCap-10M}

\begin{table}[!t]
\centering
\resizebox{1.0\linewidth}{!}{
\begin{tabular}{lrrc}
\toprule
Dataset                                             & Images        & Objects       & Avg Caption Length \\
\midrule
Flickr Entities~\cite{flickr30k}                    & 31,783        & 275,775       & - \\
RefCOCOg~\cite{refcocog}                            & 26,711        & 54,822        & 8.43  \\
RefCOCO~\cite{refcoco}                              & 19,994        & 50,000        & 3.61  \\
RefCOCO+~\cite{refcoco}                             & 19,992        & 49,856        & 3.53  \\
Visual Genome~\cite{vg_dataset_Krishna2016VisualGC} & 108,077       & 4,102,818     & - \\
%GRIT-20M~\cite{kosmos2_abs-2306-14824}              & 20,508,818    & 36,459,659    & 3.58  \\
\midrule
RegionCap-10M (Ours)                                & 5,655,833     & 10,766,958     & 9.66  \\
\bottomrule
\end{tabular}
}
\caption{Comparison RegionCap-10M with existing open-source box-text paired datasets.}
\label{tbl:data_compare}
\end{table}

To enhance \Ours's comprehension ability at the regional level, we resort to box-text paired datasets. The box-text paired data uses a bounding box and a corresponding caption to depict each instance or region in an image, e.g., RefCOCO~\cite{refcoco}. However, existing box-text paired datasets are relatively scarce due to their expensive and time-consuming annotations. 
In this work, we also construct a web-scale box-text paired dataset, \textbf{RegionCap-10M}, which is built upon a subset of large-scale image datasets including CC3M~\cite{cc3m_SoricutDSG18}, C12M~\cite{cc12m_ChangpinyoSDS21}, and COYO-700M~\cite{coyo700m}. 
As shown in Table~\ref{tbl:data_compare}, we compare RegionCap-10M with existing publicly accessible box-text paired datasets. 
A comprehensive delineation of the data construction process can be found in the appendix.
We will open-source RegionCap-10M to the community, and take experiments on the large-scale RegionCap-10M dataset in our future work.

%% file: chapters/5_conclusion.tex
\section{Conclusion}
This work proposes a unified MLLM framework named \Ours{} that integrates holistic and regional object comprehension. \Ours{} presents a PaFE module that can efficiently extract text-aligned regional features from regular image features and irregular point cloud features to support comprehending regional objects. To efficiently extend the modalities comprehended by the existing pre-trained BLIP-2, \Ours{} freezes the Q-Former and learns a set of modality-specific Lora parameters for the newly added point cloud and regional object modalities.

%% file: chapters/region_text_data.tex
%\section{RegionCap Dataset}
%\label{sec:box_text_dataset}

\subsection{Construction Process of RegionCap-10M}

\begin{figure*}[!t]
    \centering
    \includegraphics[width=0.96\textwidth]{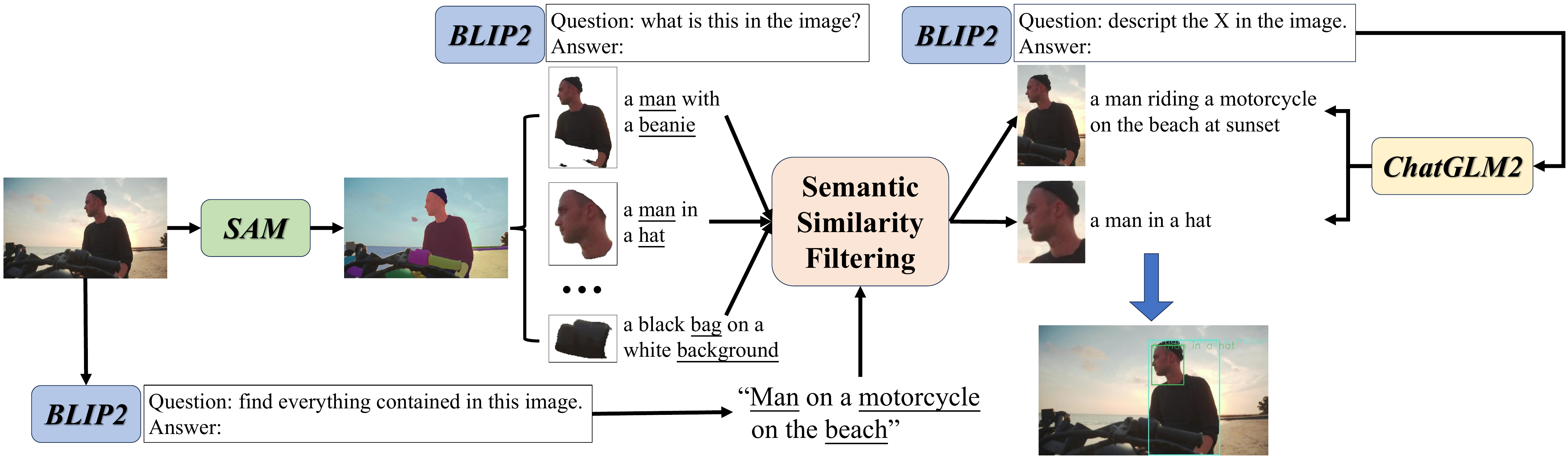}
    \caption{Illustration of our automated pipeline for extracting box-text paired data from large-scale image datasets.}
    \label{fig:regioncap}
    \vspace{-.1in}
\end{figure*}

\begin{figure*}[!t]
    \centering
    \includegraphics[width=0.96\linewidth]{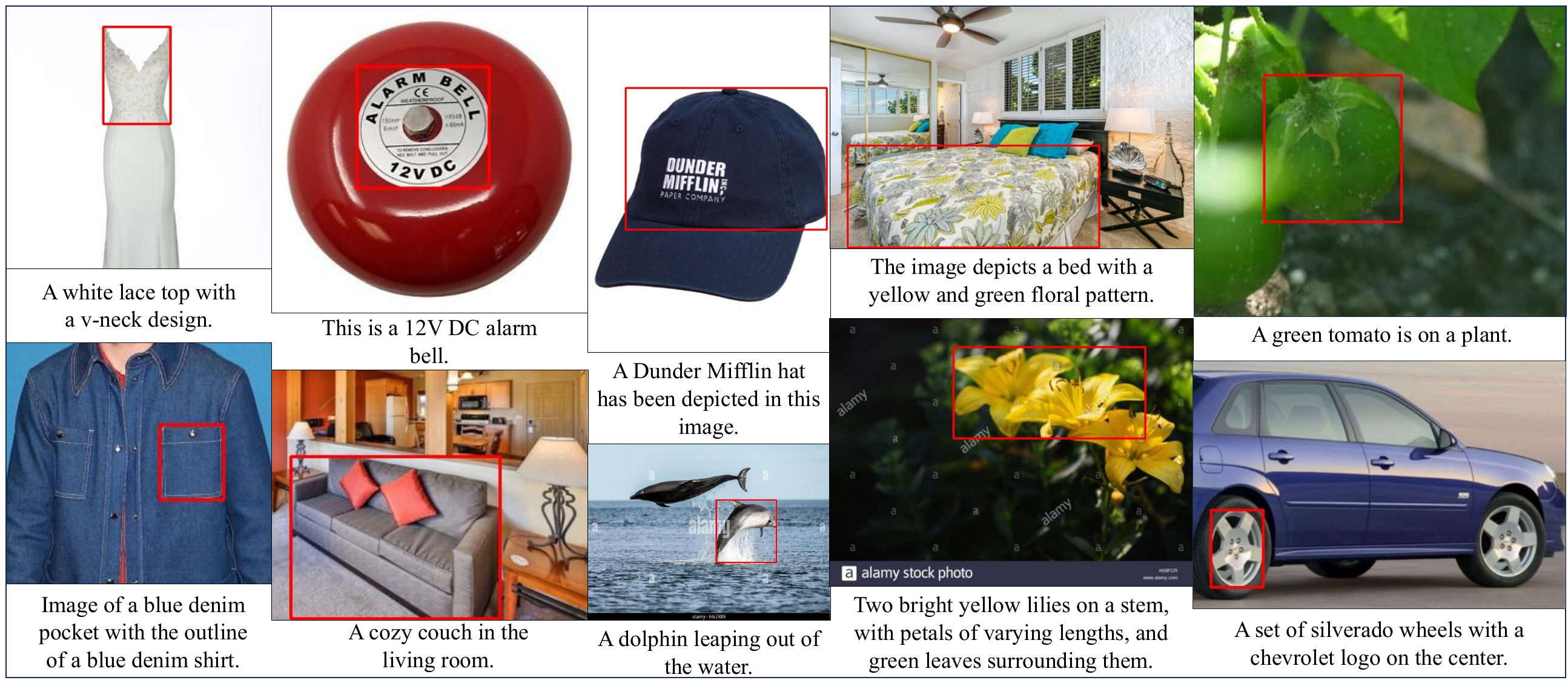}
    \vspace{-.1in}
    \caption{Visualization of data samples of RegionCap-10M.}
    \label{fig:regioncap_vis}
    \vspace{-.1in}
\end{figure*}

Data samples of RegionCap-10M are shown in Figure~\ref{fig:regioncap_vis}. We use 15$\times$8 V100 16G GPUs and spend 14 days to construct our RegionCap-10M dataset.
As depicted in Figure~\ref{fig:regioncap}, the pipeline mainly consists of three steps: class-agnostic region discovery, regional caption generation, and regional caption refinement. We describe each step in detail as follows:

\paragraph{Step-1: Class-agnostic region discovery.}
Given an image, we aim to extract each region or instance and generate a corresponding caption. We first extract as many regions or instances as possible based on a powerful segmentation model, Segment Anything Model (SAM)~\cite{sam}. SAM is pre-trained based on a large-scale segmentation dataset SA-1B and possesses strong zero-shot generalization ability at part segmentation, which enables us to extract as rich regions as possible. In our work, we use SAM to automatically generate object masks for the whole image using a grid of points as the prompt.

\paragraph{Step-2: Regional caption generation.}
Once we get the segmentation mask for each region, we can extract each region from the image and use a pre-trained image caption model to generate a corresponding description. In this work, we use the pre-trained BLIP-2~\cite{blip2_2023} as the image caption model. As an image-text pre-training model, BLIP-2 based on LLM possesses powerful zero-shot image-to-text generation ability. We find that the generated regional captions can be easily affected by the background, which is also revealed by CAT~\cite{captionanything}. Thus in this step, we replace the background with white for each cropped regional image (as depicted in Figure~\ref{fig:regioncap}), and ask BLIP-2 to identify the thing in the regional image.

\paragraph{Step-3: Regional caption refinement.}
In this step, we filter out irrelevant captions and refine the relevant captions to make them more consistent with the image.
First, we propose a semantic similarity filtering module to filter out box-texts pairs irrelevant to the image caption. Specifically, given a regional caption and the image caption, we use spaCy~\cite{spacy} to parse the captions and extract all noun chunks, denoted as $\{ n^i_{I}|^{N_I}_{i=1} \}$ and $\{ n^i_{R}|^{N_R}_{i=1} \}$ respectively. We then use CLIP's text encoder $E_t$ to calculate the semantic similarity between these two noun sets and filter out captions with the max similarity less than $\tau$ (0.9 by default):
\begin{equation}
\begin{cases}
\text{retain}, & \max_{i\in [1, N_I], j \in [1, N_R]} E_t(n^i_I) \cdot E_t(n^j_R) > \tau, \\ 
\text{filter out}, & \text{otherwise},
\end{cases} 
\end{equation}

Secondly, for each retained box-text pair, we crop the region with a background and ask BLIP-2 to describe the $X$ in the image, where $X$ denotes the identified name from step-2. In this way, BLIP-2 can accurately describe the regions of interest in the image. Finally, we leverage ChatGPT-like LLMs to convert the regional captions into compact and coherent sentences and filter out duplicate and non-English sentences. In this work, we use the open-source ChatGLM2-6B~\cite{chatglm}.
We visualize some data samples of RegionCap-10M in Figure~\ref{fig:regioncap_vis}. Our extraction method can automatically generate reasonable box regions and corresponding captions for each image.

\begin{comment}
    
\subsection{Experiment with RegionCap-10M Dataset}
We randomly select 10k paired box-text data from the collected RegionCap-10M as the test set and the rest as the training set for the experiments. As shown in Table~\ref{tbl:exp_regioncap_10m}, on the test set of RegionCap-10M, the pre-trained ReginBLIP model obtained a CIDEr score of 206.2. However, when testing the pre-trained model on RefCOCO’s test set in a zero-shot fashion, we can only achieve a CIDEr score 18.1. We will further increase the size of the RegionCap dataset to improve the generalization of image-region comprehension for MLLM models trained on it.

\begin{table}[!t]
\centering
\resizebox{0.6\linewidth}{!}{
\begin{tabular}{@{}cc|cc@{}}
\toprule
\multicolumn{2}{c|}{Zero-shot RefCOCO (test)} & \multicolumn{2}{l}{RegionCap (test)} \\
\multicolumn{1}{c}{CIDEr}       & SPICE      & \multicolumn{1}{c}{CIDEr}  & SPICE  \\ \midrule
\multicolumn{1}{c}{18.5}            &       14.6     & \multicolumn{1}{c}{206.2}       &   33.8     \\
\bottomrule
\end{tabular}
}
\caption{Zero-shot performance on the RefCOCO test set. The model was pre-trained on our collected RegionCap-10M dataset.
The RegionBLIP OPT$_{2.7B}$ model was employed in this study}
\label{tbl:exp_regioncap_10m}
\end{table}

\end{comment}